\pdfoutput=1

\documentclass[11pt]{article}

\usepackage[final]{acl}

\usepackage{times}
\usepackage{latexsym}

\usepackage[T1]{fontenc}
\usepackage[utf8]{inputenc}
\usepackage{microtype}
\usepackage{inconsolata}
\usepackage{graphicx}

\usepackage{booktabs}
\usepackage{amssymb}
\usepackage{multirow}
\usepackage{amsmath}

\title{SAGE: Strategy-Adaptive Generation Engine for Query Rewriting}

\author{
Teng Wang \\
OPPO Research Institution \\
\texttt{wt0318@connect.hku.hk}
\And
Hailei Gong \\
Tsinghua University \\
\texttt{gonghl18@mails.tsinghua.edu.cn}
\AND
Changwang Zhang \\
OPPO Research Institution \\
\texttt{zhangchangwang@oppo.com}
\And
Jun Wang \\
OPPO Research Institution \\
\texttt{wangjun7@oppo.com}
}

\begin{document}
\maketitle
\begin{abstract}

Query rewriting is pivotal for enhancing dense retrieval, yet current methods demand large-scale supervised data or suffer from inefficient reinforcement learning (RL) exploration. 
In this work, we first establish that guiding Large Language Models (LLMs) with a concise set of expert-crafted strategies substantially improves retrieval effectiveness on challenging benchmarks, including HotpotQA, FEVER, NFCorpus, and SciFact. 
Building on this insight, we introduce the \textbf{S}trategy-\textbf{A}daptive \textbf{G}eneration \textbf{E}ngine (\textbf{SAGE}), which operationalizes these strategies in an RL framework.
SAGE introduces two novel reward shaping mechanisms- \textbf{S}trategic \textbf{C}redit \textbf{S}haping (\textbf{SCS}) and \textbf{C}ontrastive \textbf{R}eward \textbf{S}haping (\textbf{CRS})-to deliver more informative learning signals. 
This strategy-guided approach not only achieves new state-of-the-art NDCG@10 results, but also uncovers a compelling emergent behavior: 
the agent learns to select optimal strategies, reduces unnecessary exploration, and generates concise rewrites, lowering inference cost without sacrificing performance.
Our findings demonstrate that strategy-guided RL, enhanced with nuanced reward shaping, offers a scalable, efficient, and more interpretable paradigm for developing the next generation of robust information retrieval systems.

\end{abstract}

\section{Introduction}

Effective information retrieval (IR) is increasingly reliant on dense retrieval systems, which map queries and documents into a shared semantic space. The performance of these systems, however, is fundamentally bound by the quality of the input query. 
To bridge the significant gap between a user's initial intent and a query optimized for machine comprehension, query rewriting has emerged as a critical component. 
While LLMs have shown significant promise for data generation, comprehension, and reasoning~\cite{gpt4,mlprompt,qwen2.5,qwen3,HRM}, current methodologies face two primary obstacles: traditional supervised fine-tuning demands large-scale, costly manual annotations, whereas modern RL approaches, such as PPO~\cite{ppo} and GRPO~\cite{grpo}, often struggle with inefficient exploration. This inefficiency not only hampers the discovery of optimal rewriting strategies but can also result in unstable training dynamics and even catastrophic failures, where models produce incoherent or irrelevant outputs.

While prior work by \citet{DMQR-RAG} has explored strategy-based prompting, we find their strategies, specifically designed for sparse retrieval in general web search, struggle to generalize to the nuanced demands of dense retrieval and exhibit limited effectiveness on specialized benchmarks.
%
To address these challenges, we propose five novel query-rewriting strategies specifically designed for dense retrieval scenarios. These strategies significantly enhance the ability of LLMs to effectively reformulate queries, consistently outperforming previous method~\cite{DMQR-RAG} on diverse and challenging benchmarks, including HotpotQA~\cite{hotpotqa}, FEVER~\cite{fever}, NFCorpus~\cite{nfcorpus}, and SciFact~\cite{scifact}. 

\begin{figure*}[t]
\centering
\includegraphics[width=0.9\textwidth]{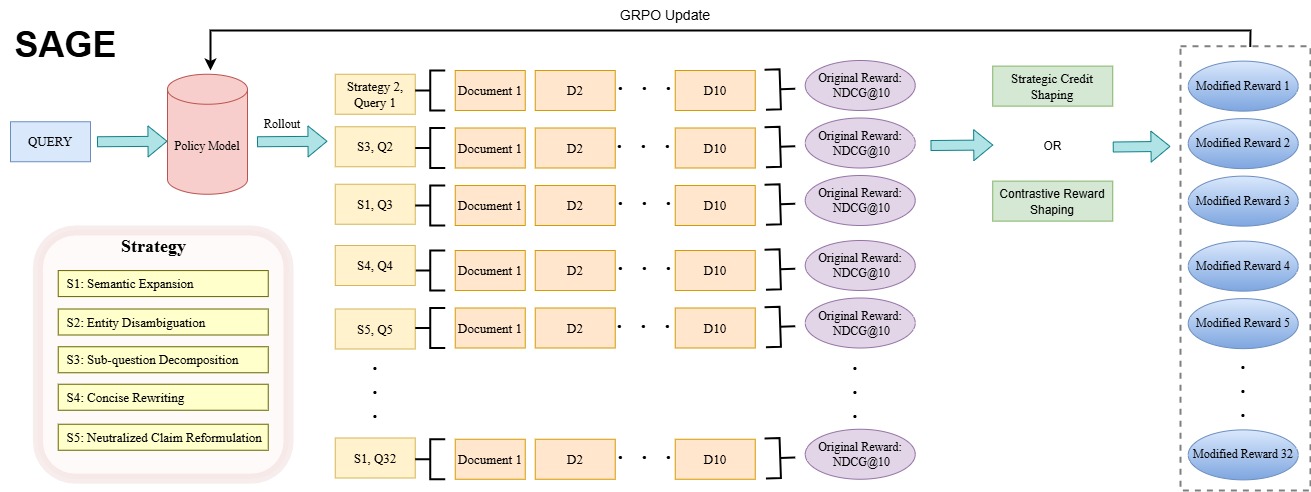}
\caption{An overview of the \textbf{SAGE} framework. SAGE operationalizes expert-crafted strategies within a reinforcement learning loop. The policy model generates a strategy-guided rewrite, which is evaluated against the environment to produce an initial reward (NDCG@10). This reward is then transformed by our novel shaping modules-\textbf{S}trategic \textbf{C}redit \textbf{S}haping or \textbf{C}ontrastive \textbf{R}eward \textbf{S}haping-to create a potent learning signal for the agent.}
\label{fig:sage}
\end{figure*}

The successful application of on-policy RL to fine-tune LLMs is pioneered by algorithms such as PPO~\cite{ppo}, which in turn inspired a family of variants including GRPO~\cite{grpo}, VAPO~\cite{VAPO}, and DAPO~\cite{DAPO}. While these powerful algorithms have enabled new capabilities in query rewriting for LLMs, many prior approaches~\cite{deep_retrieval} simply adopt them as generic optimization tools, without incorporating domain-specific guidance tailored to the query rewriting task.

To bridge this gap, we introduce the \textbf{S}trategy-\textbf{A}daptive \textbf{G}eneration \textbf{E}ngine \textbf{(SAGE)}, as demonstrated in Fig.~\ref{fig:sage}, a framework that directly integrates our human-designed strategies into the GRPO algorithm, steering LLMs toward more effective query-rewriting policies. 
%
To further refine the learning signal, we propose two novel reward shaping schemes. In addition to using the direct NDCG@10 score, we introduce \textbf{S}trategic \textbf{C}redit \textbf{S}haping (\textbf{SCS}), which assigns credit based on the average performance of each strategy, and \textbf{C}ontrastive \textbf{R}eward \textbf{S}haping (\textbf{CRS}), which transforms absolute scores into relative performance measures.
%


By reproducing prior approaches~\cite{DMQR-RAG,deep_retrieval}, we identify a significant form of reward hacking~\cite{lilianweng,reward_hacking_in_llm1} inherent in the query rewriting task.
When fine-tuned on datasets such as HotpotQA~\cite{hotpotqa}, models tend to converge to the trivial policy of directly copying the input query provided in the prompt, as modern retrievers like BGE-en-v1.5~\cite{bge_embedding} already achieve strong baseline performance with the original queries. 
Notably, even semantically appropriate rewrites frequently result in lower retrieval scores, discouraging exploration and trapping the agent in a local optimum that precludes the discovery of potentially superior strategies.

%

To address this issue, we promote exploration by modifying the prompt and penalizing outputs identical to the original query. This approach encourages the agent to venture beyond the safe default. The effectiveness of this strategy is further validated by detailed ablation studies in Section~\ref{sec:ablation_study:exploration}.

We evaluate SAGE on two challenging datasets: HotpotQA~\cite{hotpotqa} and NFCorpus~\cite{nfcorpus}. Our results demonstrate that SAGE achieves state-of-the-art retrieval effectiveness as measured by NDCG@10. Notably, we observe an emergent behavior where SAGE learns a more efficient reasoning process, substantially reducing token usage. This improved efficiency directly translates to lower inference latency and reduced computational costs.

Our main contributions are summarized as follows:
\begin{enumerate}
\item We show that prompting LLMs with a small set of interpretable strategies substantially improves query rewriting quality, establishing the performance upper bound attainable through prompting alone.

\item We introduce \textbf{SAGE}, a novel RL framework that systematically integrates explicit strategies into the learning process. SAGE autonomously adapts strategy selection via our proposed reward-shaping mechanisms, enabling more effective policy optimization.

\item We establish a new state-of-the-art in dense retrieval effectiveness (NDCG@10) using SAGE, while identifying a notable emergent behavior: SAGE learns a more efficient reasoning process, substantially reducing inference latency and computational costs.

\item We provide comprehensive analyses and ablation studies highlighting the critical importance of forced exploration. Our results underscore the necessity of an explicit penalty mechanism to avoid reward hacking, offering valuable insights for effectively training RL-based rewriting models.
\end{enumerate}
\section{Methodology}


The core idea of our methodology is to replace black-box RL optimization with explicit, human-interpretable decision-making. To this end, we propose the \textbf{S}trategy-\textbf{A}daptive \textbf{G}eneration \textbf{E}ngine (\textbf{SAGE}), which reformulates query rewriting as a structured, strategy-driven decision process. Our framework directly incorporates expert-crafted strategies into RL training to systematically guide both exploration and policy learning, as illustrated in Figure~\ref{fig:sage}.




\subsection{Problem Formulation}

We frame query rewriting as an RL task. Given an initial user query $q_{orig}$ and a document collection $\mathcal{D}$, the objective is to learn a policy $\pi$ that generates an improved query $q$. The environment evaluates the rewritten query using NDCG@10, yielding an initial reward $r_{\text{orig}} = \text{NDCG@10}(q, \mathcal{D})$, which is further refined by our reward shaping methods (see Section~\ref{sec:reward_shaping}) to produce the final reward $r_{\text{final}}$. The agent thus aims to maximize the expected final reward: $\mathbb{E}_{q \sim \pi(\cdot|q_{orig})}[r_{\text{final}}]$. We use GRPO~\cite{grpo} as our optimization algorithm and incorporate a set of expert-crafted rewriting strategies $\mathcal{S}=\{s_1, s_2, \dots, s_5\}$, providing explicit guidance for query reformulation.

\subsection{Explicit Strategic Primitives}
\label{sec:method:strategy}

To move beyond generic prompting, we introduce five explicit, expert-crafted query rewriting strategies, each tailored to address specific challenges in dense retrieval. These strategies systematically alter the semantic structure of queries to mitigate common retrieval failures.
The primitives, summarized in Table~\ref{tab:strategies}, are not mutually exclusive and often involve inherent trade-offs (e.g., semantic expansion may improve recall at the expense of precision). Collectively, they constitute the discrete action space from which our SAGE framework dynamically selects and applies. Detailed prompt templates for each strategy are provided in Appendix~\ref{appendix:strategy}.

\subsection{The SAGE Framework}

\textbf{SAGE} operationalizes our expert-crafted strategies by embedding them within an on-policy RL algorithm. In contrast to black-box methods, SAGE transforms the task from unconstrained text generation into a structured, two-part action selection process, making the agent's decision-making more explicit and interpretable. The overall workflow is illustrated in Figure~\ref{fig:sage}.

For each input query, the policy model generates a batch of $N$ rollouts, where each action is a structured pair, $\{q_i, s_i\}$, comprising the rewritten query and the integer ID of the selected strategy. This formulation requires the LLM to produce not only an effective rewrite but also an explicit strategy choice, rendering its intent interpretable.

Each rewritten query $q_i$ is evaluated by the retrieval environment to obtain an initial reward $r_{\text{orig}, i}$, measured by NDCG@10.
%
In conventional RL, this raw score serves as the final reward; however, we argue that it provides insufficient learning signal, as it neither credits effective strategy selection nor clearly distinguishes superior rewrites within a batch.
To address this, we process the initial rewards with a dedicated reward shaping module (see Section~\ref{sec:reward_shaping}), which transforms the raw scores into a more informative signal $r_{\text{final}, i}$ for policy updates.

\begin{figure*}[t]
\centering
\includegraphics[width=0.9\textwidth]{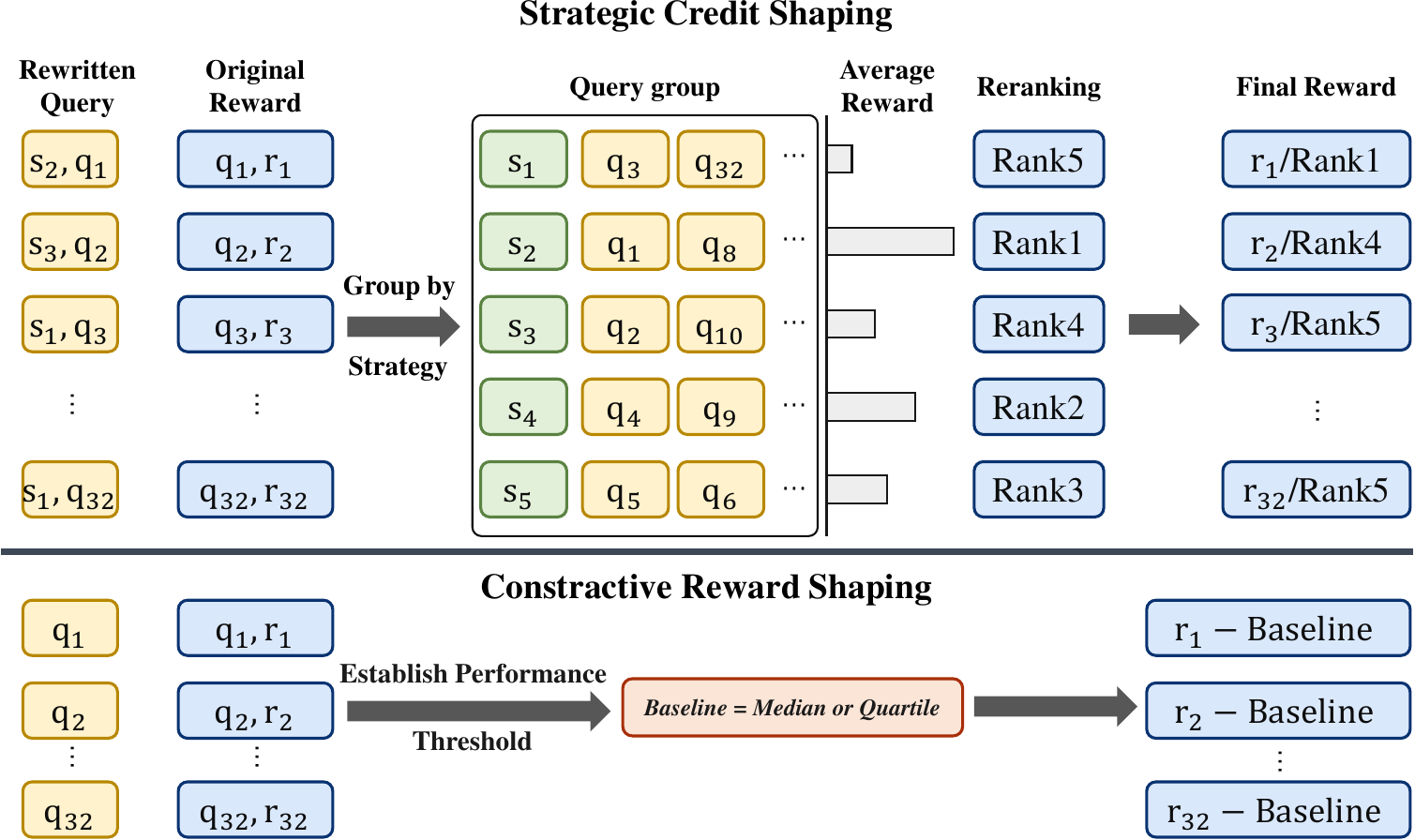}

\caption{An illustration of our two proposed reward shaping mechanisms. (a) SCS solves the credit assignment problem by grouping rollouts based on their chosen strategy, ranking these strategies by their average performance, and then scaling the individual rewards by their strategy's rank. (b) CRS sharpens the learning signal by normalizing each reward against a baseline (e.g., the batch median), reframing the objective as outperforming the typical performance.}
\label{fig:SCS_CRS}
\end{figure*}

\subsection{Reward Shaping Mechanisms}
\label{sec:reward_shaping}

To provide the agent with a more informative learning signal, we introduce two novel reward shaping mechanisms that go beyond simply using the raw NDCG@10 score. While the direct score is a standard choice, it provides only limited feedback: it fails to explicitly reward the selection of high-performing strategies and does not encourage intra-batch competition, as it treats all positive outcomes similarly-making little distinction between adequate and exceptional rewrites. To address these shortcomings, we propose two new mechanisms that transform the raw score into a more effective learning objective, as illustrated in Figure~\ref{fig:SCS_CRS}.


\paragraph{Strategic Credit Shaping (SCS):} This method aims to explicitly solve the credit assignment problem by rewarding the agent for selecting high-performing strategies. Within a given batch of rollouts, we first group them by their chosen strategy $s_i$. We then compute the average initial reward, $\bar{r}_{\text{orig}}(s_i)$, for each strategy group and rank them based on this score. The final reward for an individual rollout is its original score scaled by the inverse of its strategy's rank:
    \begin{equation}
        r_{\text{SCS}, i} = \frac{r_{\text{orig}, i}}{\text{rank}(s_i)}
        \label{eq:scs}
    \end{equation}
    This mechanism directly encourages the agent to converge on strategies that are collectively more effective, providing a clearer signal for strategic decision-making.

\paragraph{Contrastive Reward Shaping (CRS):} This method introduces intra-batch competition by normalizing rewards against a dynamic baseline, effectively penalizing underperforming rewrites while rewarding those that surpass the batch's typical performance. The final reward is the advantage over this baseline:
    \begin{equation}
        r_{\text{CRS}, i} = r_{\text{orig}, i} - \text{baseline}
        \label{eq:crs}
    \end{equation}
This forces the agent to learn policies that outperform the batch's typical performance, rather than simply achieving any positive score. By creating a clearer distinction between superior and mediocre rewrites, CRS sharpens the reward landscape and accelerates learning.

\subsection{Countering Reward Hacking with Forced Exploration}
\label{sec:reward_hacking}

Reproducing previous work~\cite{deep_retrieval,DMQR-RAG}, we observe pervasive reward hacking in query rewriting: agents exploit reward function loopholes to achieve high scores without fulfilling the intended objective. In our setting, the strong baseline performance of modern retrievers allows agents to obtain high rewards simply by copying the original query, a strategy the policy model quickly adopts since it has direct access to the input. As a result, the agent becomes trapped in a deceptive local optimum, stifling exploration of potentially superior rewriting strategies.
To directly counteract this, we introduce two mechanisms designed to force exploration.

\paragraph{Exploration Penalty.}

We apply a simple penalty term to disincentivize the agent from reverting to the trivial policy of copying the input. The final reward, $r_{\text{final}}$, is calculated by applying this penalty to the base reward signal, $r_{\text{base}}$, which represents the output from our main reward calculation (either direct NDCG@10, SCS, or CRS). The relationship is formally defined as:

\begin{equation}
\label{eq:penalty}
r_{\text{final}} =
\begin{cases}
r_{\text{base}} - p & \text{if } q_{orig} = q \\
r_{\text{base}}      & \text{otherwise}
\end{cases}
\end{equation}

where $q_{orig}$ is the original query, $q$ is the rewritten query, and $p$ is a fixed penalty hyperparameter. This formulation ensures that the agent is explicitly penalized only when it outputs a query identical to the input, thereby directly encouraging the exploration of novel rewrites.

\begin{figure*}[t]
\centering
\includegraphics[height=0.5\textwidth]{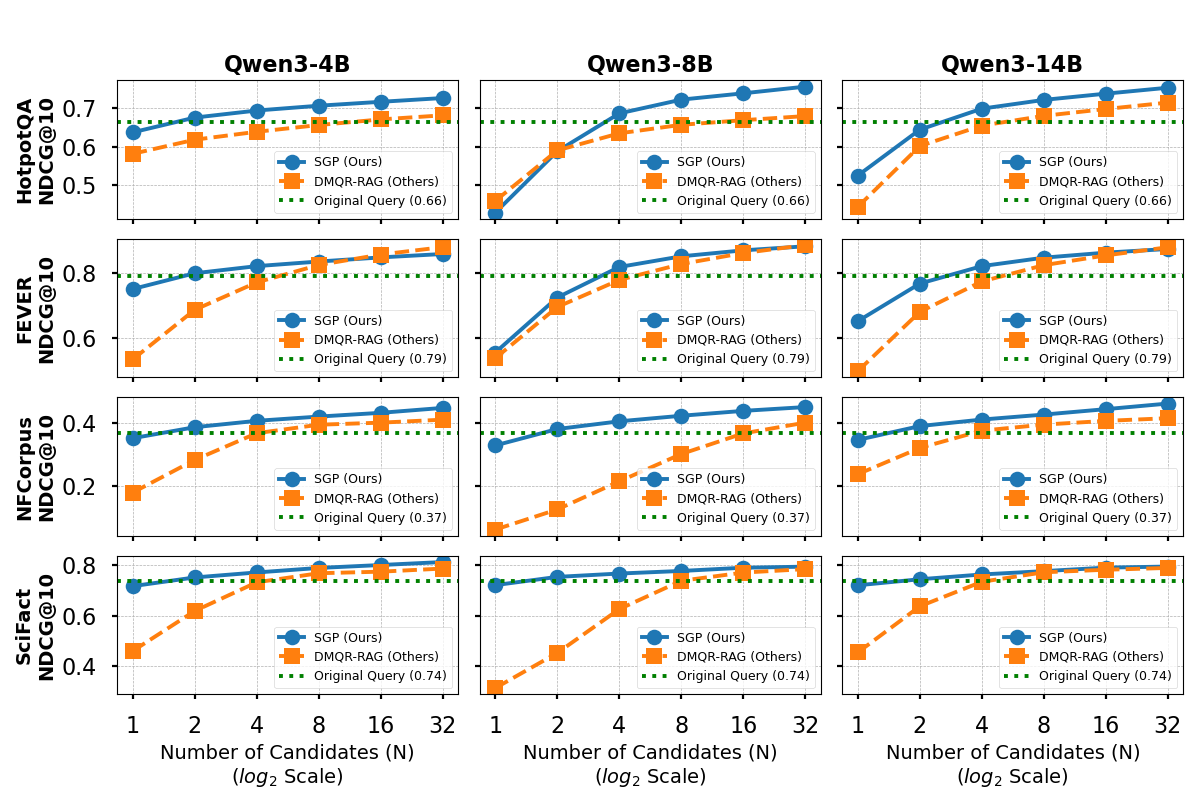}

\caption{Performance scaling laws for query rewriting using the BGE-base-en-v1.5 retriever across four benchmark datasets: HotpotQA, FEVER, NFCorpus, and SciFact. Each subplot compares the upper-bound performance (NDCG@10) of our Strategy-Guided Prompting (SGP) against the strategies from DMQR-RAG\cite{DMQR-RAG}, evaluated via a Best-of-N methodology across the Qwen3 model series~\cite{qwen3}. 
}
\label{fig:bge_scaling_law}
\end{figure*}

\paragraph{Proactive Exploration Prompting.}
In contrast to prior work~\cite{DMQR-RAG}, which instructs the model to keep the query unchanged if a better one is not found, our prompting philosophy is designed to actively encourage the agent to explore alternative rewrites.

We validate the effectiveness of these forced exploration mechanisms in our ablation studies in Section~\ref{sec:ablation_study:exploration}.
\begin{table*}[t]
\footnotesize
\centering
\renewcommand{\arraystretch}{1.2}

\begin{tabular}{l cc cc}
\toprule
\multirow{2}{*}{\textbf{Method}} & \multicolumn{2}{c}{\textbf{HotpotQA}} & \multicolumn{2}{c}{\textbf{NFCorpus}} \\
\cmidrule(lr){2-3} \cmidrule(lr){4-5}
& NDCG@10 $\uparrow$ & Avg. Tokens $\downarrow$ & NDCG@10 $\uparrow$ & Avg. Tokens $\downarrow$ \\
\midrule
\multicolumn{5}{l}{\textit{\textbf{Baselines}}} \\
\quad Original Query & 0.6633 & 0 & 0.3677 & 0 \\\hline
\quad Qwen3-4B & 0.6366 & 1598 & 0.3527 & 911 \\
\quad Qwen3-8B         & 0.4295 & 1966 & 0.3298 & 885 \\
\quad Qwen3-14B       & 0.5251 & 1682 & 0.3473 & 586 \\
\quad GPT-4.1     & \textbf{0.7118} & 297  & 0.3711 &  205\\
\quad GPT-o4-mini & 0.6915 &776& 0.3809& 440\\
\quad Claude-Sonnet-4-thinking & 0.6515& 1451 & 0.3701 & 1152 \\
\quad Gemini-2.5-Flash &0.6425& 1614 & 0.3689& 957 \\
\quad Gemini-2.5-Pro & 0.6671 & 1986 & 0.3663 & 1096 \\

\quad Deepseek-R1 & 0.6262&2182 & 0.3270 & 1251\\\hline
\quad DeepRetrieval + Qwen3-4B  & 0.6681 & 232  & 0.3676 & 343 \\
\quad DMQR-RAG + Qwen3-4B  & 0.5812 & 798 & 0.1808 &  1032 \\
\midrule
\multicolumn{5}{l}{\textit{\textbf{Our Method (SAGE)}}} \\
\quad SAGE (Direct)    & 0.6894 &  92  & 0.3776  & 229 \\
\quad SAGE-SCS         & \textbf{0.6955} & \textbf{66} & 0.3967 &  \textbf{139}\\
\quad SAGE-CRS         & 0.6918 & 69 & \textbf{0.4035} & 154 \\
\bottomrule
\end{tabular}

\caption{SAGE achieves state-of-the-art performance with remarkable efficiency. This table compares our SAGE framework, fine-tuned on Qwen3-4B, against strong baselines including the re-evaluated DeepRetrieval method and significantly larger proprietary models like Gemini-2.5-Pro and GPT-o4-mini. All models are evaluated with a maximum response length of 4096 tokens. The results demonstrate that SAGE not only outperforms other RL approaches but also achieves retrieval effectiveness (NDCG@10) competitive with these massive models, while requiring substantially fewer tokens.}

    \label{tab:SAGE}
\end{table*}

\section{Experiment}
\subsection{Effectiveness of Rewriting Strategies}
\label{sec:exp:rewriting_strategies}

To empirically validate our five expert-crafted strategies (detailed in Section~\ref{sec:method:strategy} and Appendix~\ref{appendix:strategy}), we conduct a comprehensive scaling law analysis using BGE-en-base-v1.5~\cite{bge_embedding} as the retriever. We compare our Strategy-Guided Prompting (SGP) to the strategies in DMQR-RAG~\cite{DMQR-RAG} under a Best-of-N methodology, with results shown in Figure~\ref{fig:bge_scaling_law}.
Across all four challenging datasets, including HotpotQA~\cite{hotpotqa}, FEVER~\cite{fever}, NFCorpus~\cite{nfcorpus}, and SciFact~\cite{scifact}, our SGP consistently and substantially outperforms the baseline. This demonstrates that our strategies provide a more effective upper-bound on performance by successfully guiding the LLM towards higher-quality query rewrites. Similar trends of SGP outperforming the baseline are observed when using the Contriever~\cite{contriever} retriever, with detailed results provided in Appendix~\ref{appendix:contriever_scaling_law}.


\subsection{Comparison with State-of-the-Art Baselines}

Having established the upper-bound potential of our expert-crafted strategies (Section~\ref{sec:exp:rewriting_strategies}), we now evaluate the full SAGE framework. Our preliminary analysis indicates that for datasets like FEVER and SciFact, the potential for improvement via RL is constrained, due to either a high performance baseline from the original query or a high exploration cost required for marginal gains. Consequently, to provide a more meaningful evaluation of SAGE's optimization capabilities, we focus our main RL experiments on HotpotQA and NFCorpus, which present a larger performance gap and thus a more dynamic learning environment. 

Unless otherwise specified, all experiments are conducted using BGE-en-base-v1.5. Detailed training settings are provided in Appendix~\ref{appendix:hyper}.

We benchmark SAGE against a diverse suite of strong baselines to demonstrate its effectiveness. This includes state-of-the-art open-source models Qwen3-4B, Qwen3-8B, Qwen3-14B, GPT-4.1, GPT-o4-mini, Claude-Sonnect-4, Gemini-2.5-Flash, Gemini-2.5-Pro, Deepseek-R1~\cite{qwen3,gemini,gpt4,claude,deepseek-r1}. For a fair comparison under our controlled experimental setup, we also re-evaluate prior RL-based methods, including DeepRetrieval~\cite{deep_retrieval} and the strategies from DMQR-RAG~\cite{DMQR-RAG}. Against these baselines, we report the performance of SAGE enhanced with our two novel reward shaping mechanisms, SAGE-SCS and SAGE-CRS.

The results presented in Table~\ref{tab:SAGE} clearly demonstrate the dual advantage of our SAGE framework. In terms of retrieval effectiveness, our SAGE variants establish a new state-of-the-art. Notably, SAGE, fine-tuned on Qwen3-4B, consistently outperforms specialized RL-based approaches like DeepRetrieval also fine-tuned from Qwen3-4B in our experiment setting. More strikingly, our method achieves performance that is competitive with, and in some cases superior to, significantly larger and more powerful proprietary models such as Gemini-2.5-Pro, GPT-o4-mini. 

More striking than the performance gains is the emergent efficiency of our SAGE framework. We observe that SAGE consistently generates rewrites using substantially fewer tokens than competing methods, a phenomenon that is even more pronounced when using our SCS and CRS reward shaping schemes. We attribute this to the agent learning to prioritize more direct reasoning paths, effectively discovering a shortcut to high-quality rewrites without being explicitly optimized for brevity.

This finding has profound practical implications. In a production environment, the long, static portion of our strategic prompt can be pre-processed and its KV cache stored. At inference time, the system only needs to generate a short reasoning, significantly reducing latency and computational cost. This architectural advantage, which decouples the pre-computable prompt context from the short dynamic generation, makes SAGE a highly attractive solution for real-world, performance-critical retrieval systems.


\begin{figure*}[t]
\centering
\includegraphics[width=0.9\textwidth]{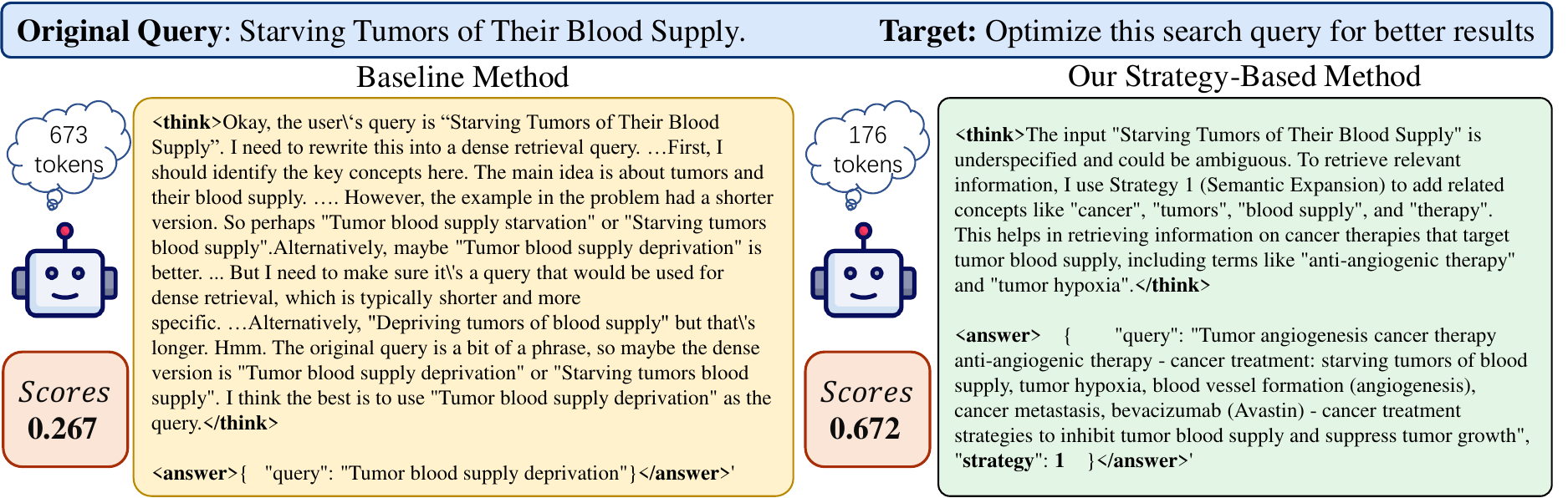}
\caption{
Qualitative comparison illustrating the mechanism behind SAGE's dual advantage. The baseline model (left), lacking strategic guidance, engages in a verbose and convoluted reasoning process, resulting in a suboptimal query. In contrast, SAGE (right) leverages an explicit strategy to find a direct and efficient path to a more semantically accurate rewrite, achieving superior retrieval effectiveness with a fraction of the token generation cost.
}
\label{fig:case_study}
\end{figure*}

\begin{table*}[t]
\centering

\begin{tabular}{l cc cc}
\toprule
\multirow{2}{*}{\textbf{Method}} & \multicolumn{2}{c}{\textbf{HotpotQA}} & \multicolumn{2}{c}{\textbf{NFCorpus}} \\
\cmidrule(lr){2-3} \cmidrule(lr){4-5}
& NDCG@10 $\uparrow$ & Avg. Tokens $\downarrow$ & NDCG@10 $\uparrow$ & Avg. Tokens $\downarrow$ \\
\midrule
\multicolumn{5}{l}{\textit{\textbf{Baseline (No Strategic Guidance)}}} \\
\quad DeepRetrieval  & 0.6681 & 232  & 0.3676 & 343 \\
\midrule
\multicolumn{5}{l}{\textit{\textbf{Our Method (with Strategic Guidance)}}} \\
\quad SAGE (Direct)    & 0.6894 &  92  & 0.3775  & 229 \\
\quad SAGE-SCS         & \textbf{0.6955} & \textbf{66} & 0.3966 &  \textbf{139}\\
\quad SAGE-CRS         & 0.6918 & 69 & \textbf{0.4035} & 154 \\
\bottomrule

\end{tabular}

\caption{Ablation study on the impact of strategic guidance. We compare the performance of a standard RL baseline against our SAGE framework, which incorporates expert-crafted strategies. The results highlight the significant gains in both effectiveness (NDCG@10) and efficiency (Avg. Tokens) brought by our strategic guidance.}
\label{tab:ablation_guidance}
\end{table*}

\section{Ablation Study}

To isolate and quantify the impact of our core contributions, we conduct a series of ablation studies. Our experiments are designed to answer several key questions: (1) What is the performance gain from our expert-crafted strategies compared to an unguided RL baseline? And what is the contribution of our novel reward shaping mechanisms (SCS and CRS) compared to a direct reward? (2) How critical is the exploration penalty for overcoming reward hacking?  

\subsection{The Impact of Strategic Guidance}
The most significant finding from our ablation study is the dramatic impact of strategic guidance. As shown in Table~\ref{tab:ablation_guidance}, removing our expert-crafted strategies and reverting to a "black-box" RL approach results in a significant degradation in performance. While this change leads to a notable drop in NDCG@10 across both datasets, the more striking effect is on efficiency, where the average response length increases dramatically. This demonstrates that models trained without strategic guidance are not only less effective but also substantially less efficient.

To provide a more intuitive understanding, Fig. ~\ref{fig:case_study} presents a direct case comparison. The output from the model trained without strategies is generic and fails to capture the query's nuance, whereas the output from SAGE is scientifically precise and far more effective for retrieval.

Furthermore, we see the additional benefits of our reward shaping mechanisms. On both datasets, SAGE-SCS and SAGE-CRS further improve upon the SAGE (Direct) baseline, pushing the performance ceiling even higher while maintaining superior efficiency. This confirms that while strategic guidance is the foundational improvement, our novel reward shaping schemes provide a crucial secondary optimization.



\begin{table*}[t]
\centering

\begin{tabular}{l c c}
\toprule
\textbf{Experimental Condition} & \textbf{NDCG@10} $\uparrow$ & \textbf{Modification Rate} $\uparrow$ \\
\midrule
 Conservative Prompt ("keep unchanged") & 0.671 & 0.509 \\
Proactive Exploration Prompt & \textbf{0.694} & 0.951 \\
Proactive Prompt + Penalty ($p=0.02$) & 0.692 & \textbf{0.998} \\
\bottomrule
\end{tabular}

\caption{Ablation study on different exploration mechanisms for SAGE-SCS on the HotpotQA dataset. We compare three settings: (1) using a conservative prompt, (2) using a proactive exploration prompt, and (3) combining the proactive prompt with an exploration penalty.}
\label{tab:ablation_exploration}
\end{table*}

\subsection{Analysis of Training Dynamics}
Beyond final performance, the training dynamics also reveal the superiority of the SAGE framework. As illustrated in Appendix~\ref{appendix:training_dynamics}, the baseline model trained without strategies quickly gets trapped in a local optimum. Its performance stagnates after only a few training steps, and its response length remains high and volatile throughout the process.

In stark contrast, SAGE demonstrates continuous performance improvement during training. More importantly, we observe a compelling emergent behavior: as SAGE learns to better utilize its strategies, it also learns to achieve the optimal rewrite with a more concise reasoning process. This results in a steady and significant decrease in the average response length over time, proving that our framework not only learns more effectively but also more efficiently.

\subsection{The Critical Role of the Exploration Penalty}
\label{sec:ablation_study:exploration}

In reproducing prior work~\cite{deep_retrieval, DMQR-RAG}, we identify a significant challenge in this task: the agent's strong tendency to revert to a trivial policy of outputting the original query without modification. We attribute this reward-hacking behavior to two primary causes. First, the high baseline performance of modern retrievers makes the "do-nothing" action a safe, high-reward option. This issue is exacerbated by instructions in prior work like \citet{DMQR-RAG}, which explicitly instructs the agent to preserve the original query if a rewrite is deemed unnecessary, reinforcing this conservative policy from the start of training. Second, many powerful retrievers have been trained on these benchmarks, making them brittle and highly sensitive to phrasing, where even minor, semantically-sound modifications can lead to a sharp drop in performance. This combination traps the agent in a deceptive local optimum, making any exploration a high-risk endeavor.

To break this cycle and force meaningful exploration, we implement a two-pronged approach. First, in contrast to prior work~\cite{DMQR-RAG} which explicitly instructs the agent to keep the original query if no better one is found, our prompt philosophy actively encourages exploration, detailed is demonstrated in Appendix~\ref{appendix:strategy}. Second, and more critically, we introduce a simple penalty term, subtracting a fixed value p=0.05 from the reward whenever the generated query is identical to the original, directly disincentivizing this reward-hacking behavior.

To validate the effectiveness of these mechanisms, we conducted an ablation study using the SAGE-SCS model on HotpotQA. The results, shown in Table~\ref{tab:ablation_exploration}, clearly demonstrate the value of our approach: simply replacing the conservative prompt with our proactive exploration prompt substantially increases both NDCG@10 and the modification rate, highlighting the critical role of encouraging exploration.
However, the use of an exploration penalty reveals a critical trade-off. While it drives the modification rate to 0.998, we observe a slight decrease in NDCG@10. This suggests that in cases where the original query is already optimal, enforcing modifications via a penalty may slightly impair peak performance. Thus, although the penalty is highly effective for maximizing exploration and mitigating reward hacking, its application should be carefully balanced against the specific task’s need for exhaustive exploration versus optimal end performance.

\section{Conclusion}
In this work, we introduced SAGE, a novel framework designed to address the black-box nature and inefficient exploration of conventional RL-based query rewriting. By integrating expert-crafted strategies and novel reward shaping mechanisms (SCS and CRS) into the RL loop, SAGE achieves state-of-the-art retrieval effectiveness and emergent efficiency, substantially reducing inference costs. Importantly, our approach effectively mitigates reward hacking by explicitly promoting exploration and penalizing trivial, unmodified rewrites. These findings demonstrate that strategy-guided RL not only enhances effectiveness and efficiency but also leads to more transparent and controllable retrieval systems.

\section{Limitation}
\subsection{Evaluation Datasets}

One of the key innovations introduced in this paper is the query rewrite strategy or policy. This approach relies on manual analysis to craft rewriting strategies tailored to specific datasets. However, this methodology poses certain limitations. In particular, it is less applicable to domains such as sparse retrieval and SQL-based retrieval, where the query structure is already well-defined and leaves little room for rewriting strategies. Consequently, the benefits of our approach are constrained in these scenarios. Our work is therefore primarily focused on the dense retrieval setting, where more flexibility in query reformulation exists.

\bibliography{custom}

\begin{thebibliography}{37}
\providecommand{\natexlab}[1]{#1}

\bibitem[{Abdul-Jaleel et~al.(2004)Abdul-Jaleel, Allan, Croft, Diaz, Larkey, Li, Smucker, and Wade}]{traditional_query_aug2}
Nasreen Abdul-Jaleel, James Allan, W~Bruce Croft, Fernando Diaz, Leah Larkey, Xiaoyan Li, Mark~D Smucker, and Courtney Wade. 2004.
\newblock Umass at trec 2004: Novelty and hard.
\newblock In \emph{Proceedings of TREC-13}, pages 715--725.

\bibitem[{Achiam et~al.(2023)Achiam, Adler, Agarwal, Ahmad, Akkaya, Aleman, Almeida, Altenschmidt, Altman, Anadkat et~al.}]{gpt4}
Josh Achiam, Steven Adler, Sandhini Agarwal, Lama Ahmad, Ilge Akkaya, Florencia~Leoni Aleman, Diogo Almeida, Janko Altenschmidt, Sam Altman, Shyamal Anadkat, and 1 others. 2023.
\newblock Gpt-4 technical report.
\newblock \emph{arXiv preprint arXiv:2303.08774}.

\bibitem[{Anthropic(2025)}]{claude}
Anthropic. 2025.
\newblock \href {https://www-cdn.anthropic.com/de8ba9b01c9ab7cbabf5c33b80b7bbc618857627/Model_Card_Claude_3.pdf} {The claude 3 model family: Opus, sonnet, haiku}.

\bibitem[{Boteva et~al.(2016)Boteva, Gholipour, Sokolov, and Riezler}]{nfcorpus}
Vera Boteva, Demian Gholipour, Artem Sokolov, and Stefan Riezler. 2016.
\newblock A full-text learning to rank dataset for medical information retrieval.
\newblock In \emph{Advances in Information Retrieval: 38th European Conference on IR Research, ECIR 2016, Padua, Italy, March 20--23, 2016. Proceedings 38}, pages 716--722. Springer.

\bibitem[{Dalton et~al.(2014)Dalton, Dietz, and Allan}]{traditional_query_aug3}
Jeffrey Dalton, Laura Dietz, and James Allan. 2014.
\newblock Entity query feature expansion using knowledge base links.
\newblock In \emph{Proceedings of the 37th international ACM SIGIR conference on Research \& development in information retrieval}, pages 365--374.

\bibitem[{Devlin et~al.(2018)Devlin, Chang, Lee, and Toutanova}]{bert}
Jacob Devlin, Ming-Wei Chang, Kenton Lee, and Kristina Toutanova. 2018.
\newblock Bert: Pre-training of deep bidirectional transformers for language understanding.
\newblock \emph{arXiv preprint arXiv:1810.04805}.

\bibitem[{Guo et~al.(2025)Guo, Yang, Zhang, Song, Zhang, Xu, Zhu, Ma, Wang, Bi et~al.}]{deepseek-r1}
Daya Guo, Dejian Yang, Haowei Zhang, Junxiao Song, Ruoyu Zhang, Runxin Xu, Qihao Zhu, Shirong Ma, Peiyi Wang, Xiao Bi, and 1 others. 2025.
\newblock Deepseek-r1: Incentivizing reasoning capability in llms via reinforcement learning.
\newblock \emph{arXiv preprint arXiv:2501.12948}.

\bibitem[{Izacard et~al.(2021)Izacard, Caron, Hosseini, Riedel, Bojanowski, Joulin, and Grave}]{contriever}
Gautier Izacard, Mathilde Caron, Lucas Hosseini, Sebastian Riedel, Piotr Bojanowski, Armand Joulin, and Edouard Grave. 2021.
\newblock Unsupervised dense information retrieval with contrastive learning.
\newblock \emph{arXiv preprint arXiv:2112.09118}.

\bibitem[{Jiang et~al.(2025)Jiang, Lin, Cao, Tian, Kang, Wang, Sun, and Han}]{deep_retrieval}
Pengcheng Jiang, Jiacheng Lin, Lang Cao, Runchu Tian, SeongKu Kang, Zifeng Wang, Jimeng Sun, and Jiawei Han. 2025.
\newblock \href {https://arxiv.org/abs/2503.00223} {Deepretrieval: Hacking real search engines and retrievers with large language models via reinforcement learning}.
\newblock \emph{ArXiv e-prints}.

\bibitem[{Lewis et~al.(2020)Lewis, Liu, Goyal, Ghazvininejad, Mohamed, Levy, Stoyanov, and Zettlemoyer}]{bart}
Mike Lewis, Yinhan Liu, Naman Goyal, Marjan Ghazvininejad, Abdelrahman Mohamed, Omer Levy, Veselin Stoyanov, and Luke Zettlemoyer. 2020.
\newblock \href {https://doi.org/10.18653/v1/2020.acl-main.703} {{BART}: Denoising sequence-to-sequence pre-training for natural language generation, translation, and comprehension}.
\newblock In \emph{Proceedings of the 58th Annual Meeting of the Association for Computational Linguistics}, pages 7871--7880, Online. Association for Computational Linguistics.

\bibitem[{Li et~al.(2024)Li, Wang, Mao, Jiang, Chen, Jiazhen, Zhang, ZHANG, and Liu}]{DMQR-RAG}
Zhicong Li, Jiahao Wang, Hangyu Mao, ZhiShu Jiang, Zhongxia Chen, Du~Jiazhen, Fuzheng Zhang, Di~ZHANG, and Yong Liu. 2024.
\newblock \href {https://arxiv.org/abs/2411.13154} {Dmqr-rag: Diverse multi-query rewriting in retrieval-augmented generation}.
\newblock \emph{ArXiv e-prints}.

\bibitem[{Liu et~al.(2019)Liu, Ott, Goyal, Du, Joshi, Chen, Levy, Lewis, Zettlemoyer, and Stoyanov}]{roberta}
Yinhan Liu, Myle Ott, Naman Goyal, Jingfei Du, Mandar Joshi, Danqi Chen, Omer Levy, Mike Lewis, Luke Zettlemoyer, and Veselin Stoyanov. 2019.
\newblock Roberta: A robustly optimized bert pretraining approach.
\newblock \emph{arXiv preprint arXiv:1907.11692}.

\bibitem[{Raffel et~al.(2020)Raffel, Shazeer, Roberts, Lee, Narang, Matena, Zhou, Li, and Liu}]{t5}
Colin Raffel, Noam Shazeer, Adam Roberts, Katherine Lee, Sharan Narang, Michael Matena, Yanqi Zhou, Wei Li, and Peter~J. Liu. 2020.
\newblock \href {http://jmlr.org/papers/v21/20-074.html} {Exploring the limits of transfer learning with a unified text-to-text transformer}.
\newblock \emph{Journal of Machine Learning Research}, 21(140):1--67.

\bibitem[{Rocchio~Jr(1971)}]{query_aug}
Joseph~John Rocchio~Jr. 1971.
\newblock Relevance feedback in information retrieval.
\newblock \emph{The SMART retrieval system: experiments in automatic document processing}.

\bibitem[{Schulman et~al.(2017)Schulman, Wolski, Dhariwal, Radford, and Klimov}]{ppo}
John Schulman, Filip Wolski, Prafulla Dhariwal, Alec Radford, and Oleg Klimov. 2017.
\newblock Proximal policy optimization algorithms.
\newblock \emph{arXiv preprint arXiv:1707.06347}.

\bibitem[{Shao et~al.(2024)Shao, Wang, Zhu, Xu, Song, Bi, Zhang, Zhang, Li, Wu et~al.}]{grpo}
Zhihong Shao, Peiyi Wang, Qihao Zhu, Runxin Xu, Junxiao Song, Xiao Bi, Haowei Zhang, Mingchuan Zhang, YK~Li, Y~Wu, and 1 others. 2024.
\newblock Deepseekmath: Pushing the limits of mathematical reasoning in open language models.
\newblock \emph{arXiv preprint arXiv:2402.03300}.

\bibitem[{Sheng et~al.(2024)Sheng, Zhang, Ye, Wu, Zhang, Zhang, Peng, Lin, and Wu}]{hybridflow}
Guangming Sheng, Chi Zhang, Zilingfeng Ye, Xibin Wu, Wang Zhang, Ru~Zhang, Yanghua Peng, Haibin Lin, and Chuan Wu. 2024.
\newblock Hybridflow: A flexible and efficient rlhf framework.
\newblock \emph{arXiv preprint arXiv: 2409.19256}.

\bibitem[{Team et~al.(2023)Team, Anil, Borgeaud, Alayrac, Yu, Soricut, Schalkwyk, Dai, Hauth, Millican et~al.}]{gemini}
Gemini Team, Rohan Anil, Sebastian Borgeaud, Jean-Baptiste Alayrac, Jiahui Yu, Radu Soricut, Johan Schalkwyk, Andrew~M Dai, Anja Hauth, Katie Millican, and 1 others. 2023.
\newblock Gemini: a family of highly capable multimodal models.
\newblock \emph{arXiv preprint arXiv:2312.11805}.

\bibitem[{Thorne et~al.(2018)Thorne, Vlachos, Christodoulopoulos, and Mittal}]{fever}
James Thorne, Andreas Vlachos, Christos Christodoulopoulos, and Arpit Mittal. 2018.
\newblock \href {https://doi.org/10.18653/v1/N18-1074} {{FEVER}: a large-scale dataset for fact extraction and {VER}ification}.
\newblock In \emph{Proceedings of the 2018 Conference of the North {A}merican Chapter of the Association for Computational Linguistics: Human Language Technologies, Volume 1 (Long Papers)}, pages 809--819, New Orleans, Louisiana. Association for Computational Linguistics.

\bibitem[{Wadden et~al.(2020)Wadden, Lin, Lo, Wang, van Zuylen, Cohan, and Hajishirzi}]{scifact}
David Wadden, Shanchuan Lin, Kyle Lo, Lucy~Lu Wang, Madeleine van Zuylen, Arman Cohan, and Hannaneh Hajishirzi. 2020.
\newblock \href {https://doi.org/10.18653/v1/2020.emnlp-main.609} {Fact or fiction: Verifying scientific claims}.
\newblock In \emph{Proceedings of the 2020 Conference on Empirical Methods in Natural Language Processing (EMNLP)}, pages 7534--7550, Online. Association for Computational Linguistics.

\bibitem[{Wang et~al.(2024{\natexlab{a}})Wang, Li, Chen, Cai, Zhu, Lin, Cao, Liu, Liu, and Sui}]{reward_hacking_in_llm1}
Peiyi Wang, Lei Li, Liang Chen, Zefan Cai, Dawei Zhu, Binghuai Lin, Yunbo Cao, Qi~Liu, Tianyu Liu, and Zhifang Sui. 2024{\natexlab{a}}.
\newblock Large language models are not fair evaluators.
\newblock In \emph{ACL}.

\bibitem[{Wang et~al.(2025{\natexlab{a}})Wang, He, Yu, Fu, and Han}]{mlprompt}
Teng Wang, Zhenqi He, Wing-Yin Yu, Xiaojin Fu, and Xiongwei Han. 2025{\natexlab{a}}.
\newblock \href {https://aclanthology.org/2025.coling-main.300/} {Large language models are good multi-lingual learners : When {LLM}s meet cross-lingual prompts}.
\newblock In \emph{Proceedings of the 31st International Conference on Computational Linguistics}, pages 4442--4456, Abu Dhabi, UAE. Association for Computational Linguistics.

\bibitem[{Wang et~al.(2025{\natexlab{b}})Wang, Jiang, He, Yang, Zheng, Li, He, Tong, and Gong}]{HRM}
Teng Wang, Zhangyi Jiang, Zhenqi He, Wenhan Yang, Yanan Zheng, Zeyu Li, Zifan He, Shenyang Tong, and Hailei Gong. 2025{\natexlab{b}}.
\newblock Towards hierarchical multi-step reward models for enhanced reasoning in large language models.
\newblock \emph{arXiv preprint arXiv:2503.13551}.

\bibitem[{Wang et~al.(2025{\natexlab{c}})Wang, Yu, He, Liu, Gong, Wu, Han, Shi, She, Zhu et~al.}]{bpp-search}
Teng Wang, Wing-Yin Yu, Zhenqi He, Zehua Liu, Hailei Gong, Han Wu, Xiongwei Han, Wei Shi, Ruifeng She, Fangzhou Zhu, and 1 others. 2025{\natexlab{c}}.
\newblock Bpp-search: Enhancing tree of thought reasoning for mathematical modeling problem solving.
\newblock \emph{ACL}.

\bibitem[{Wang et~al.(2024{\natexlab{b}})Wang, Yu, She, Yang, Chen, and Zhang}]{leverageLLMMIP}
Teng Wang, Wing-Yin Yu, Ruifeng She, Wenhan Yang, Taijie Chen, and Jianping Zhang. 2024{\natexlab{b}}.
\newblock Leveraging large language models for solving rare mip challenges.
\newblock \emph{arXiv preprint arXiv:2409.04464}.

\bibitem[{Wang et~al.(2025{\natexlab{d}})Wang, Zhang, Pang, Guo, Zheng, and Zheng}]{MaFeRw}
Yujing Wang, Hainan Zhang, Liang Pang, Binghui Guo, Hongwei Zheng, and Zhiming Zheng. 2025{\natexlab{d}}.
\newblock Maferw: Query rewriting with multi-aspect feedbacks for retrieval-augmented large language models.
\newblock In \emph{Proceedings of the AAAI Conference on Artificial Intelligence}, volume~39, pages 25434--25442.

\bibitem[{Wen et~al.(2025)Wen, Zhong, Khan, Perez, Steinhardt, Huang, Bowman, He, and Feng}]{reward_hacking_in_llm2}
Jiaxin Wen, Ruiqi Zhong, Akbir Khan, Ethan Perez, Jacob Steinhardt, Minlie Huang, Samuel~R Bowman, He~He, and Shi Feng. 2025.
\newblock Language models learn to mislead humans via rlhf.
\newblock In \emph{ICLR}.

\bibitem[{Weng(2024)}]{lilianweng}
Lilian Weng. 2024.
\newblock \href {https://lilianweng.github.io/posts/2024-11-28-reward-hacking/} {Reward hacking in reinforcement learning}.

\bibitem[{Xiao et~al.(2023)Xiao, Liu, Zhang, and Muennighoff}]{bge_embedding}
Shitao Xiao, Zheng Liu, Peitian Zhang, and Niklas Muennighoff. 2023.
\newblock \href {https://arxiv.org/abs/2309.07597} {C-pack: Packaged resources to advance general chinese embedding}.
\newblock \emph{Preprint}, arXiv:2309.07597.

\bibitem[{Xiong and Callan(2015)}]{traditional_query_aug5}
Chenyan Xiong and Jamie Callan. 2015.
\newblock Query expansion with freebase.
\newblock In \emph{Proceedings of the 2015 international conference on the theory of information retrieval}, pages 111--120.

\bibitem[{Xu et~al.(2009)Xu, Jones, and Wang}]{traditional_query_aug4}
Yang Xu, Gareth~JF Jones, and Bin Wang. 2009.
\newblock Query dependent pseudo-relevance feedback based on wikipedia.
\newblock In \emph{Proceedings of the 32nd international ACM SIGIR conference on Research and development in information retrieval}, pages 59--66.

\bibitem[{Yang et~al.(2025{\natexlab{a}})Yang, Li, Yang, Zhang, Hui, Zheng, Yu, Gao, Huang, Lv et~al.}]{qwen3}
An~Yang, Anfeng Li, Baosong Yang, Beichen Zhang, Binyuan Hui, Bo~Zheng, Bowen Yu, Chang Gao, Chengen Huang, Chenxu Lv, and 1 others. 2025{\natexlab{a}}.
\newblock Qwen3 technical report.
\newblock \emph{arXiv preprint arXiv:2505.09388}.

\bibitem[{Yang et~al.(2025{\natexlab{b}})}]{qwen2.5}
An~Yang and 1 others. 2025{\natexlab{b}}.
\newblock Qwen2.5 technical report.
\newblock \emph{arXiv preprint arXiv:2505.09388}.

\bibitem[{Yang et~al.(2018)Yang, Qi, Zhang, Bengio, Cohen, Salakhutdinov, and Manning}]{hotpotqa}
Zhilin Yang, Peng Qi, Saizheng Zhang, Yoshua Bengio, William~W Cohen, Ruslan Salakhutdinov, and Christopher~D Manning. 2018.
\newblock Hotpotqa: A dataset for diverse, explainable multi-hop question answering.
\newblock \emph{arXiv preprint arXiv:1809.09600}.

\bibitem[{Yu et~al.(2025)Yu, Zhang, Zhu, Yuan, Zuo, Yue, Fan, Liu, Liu, Liu et~al.}]{DAPO}
Qiying Yu, Zheng Zhang, Ruofei Zhu, Yufeng Yuan, Xiaochen Zuo, Yu~Yue, Tiantian Fan, Gaohong Liu, Lingjun Liu, Xin Liu, and 1 others. 2025.
\newblock Dapo: An open-source llm reinforcement learning system at scale.
\newblock \emph{arXiv preprint arXiv:2503.14476}.

\bibitem[{Yue et~al.(2025)Yue, Yuan, Yu, Zuo, Zhu, Xu, Chen, Wang, Fan, Du et~al.}]{VAPO}
Yu~Yue, Yufeng Yuan, Qiying Yu, Xiaochen Zuo, Ruofei Zhu, Wenyuan Xu, Jiaze Chen, Chengyi Wang, TianTian Fan, Zhengyin Du, and 1 others. 2025.
\newblock Vapo: Efficient and reliable reinforcement learning for advanced reasoning tasks.
\newblock \emph{arXiv preprint arXiv:2504.05118}.

\bibitem[{Zhai and Lafferty(2001)}]{traditional_query_aug1}
Chengxiang Zhai and John Lafferty. 2001.
\newblock Model-based feedback in the language modeling approach to information retrieval.
\newblock In \emph{Proceedings of the tenth international conference on Information and knowledge management}, pages 403--410.

\end{thebibliography}
\clearpage

\appendix

\begin{table*}[t]
\footnotesize
\centering
\renewcommand{\arraystretch}{1.2}

\begin{tabular}{p{0.22\textwidth} p{0.3\textwidth} p{0.33\textwidth}}
\hline
    \textbf{Strategy} & \textbf{Targeted Challenge} & \textbf{Primary Use Case } \\
    \hline
    
    Senmantic Expansion & Vocabulary mismatch between query and documents. & General purpose, especially in specialized domains (e.g., NFCorpus).\\ 
    \hline
    
    Entity Disambiguation & Ambiguous entities leading to incorrect retrieval. & Queries with common names or acronyms. \\ \hline

    Sub-question Decomposition & Complex, multi-step information needs. & Multi-hop QA (e.g., HotpotQA). \\ \hline

    Concise Rewriting & Query noise from redundant or conversational phrases. & Improving precision for overly verbose user queries. \\ \hline

Neutralized Claim Reformulation & Retriever's confirmation bias towards a stated claim. & Fact-checking and verification (e.g., FEVER, SciFact). \\
    
    \hline

\end{tabular}
 \caption{Overview of our five expert-crafted rewriting strategies and their targeted applications.}
    \label{tab:strategies}
\end{table*}

\appendix

\section{Appendix}
\subsection{Strategy-Guided Prompting}
\label{appendix:strategy}

\begin{figure*}[t]
\centering
\includegraphics[width=0.9\textwidth]{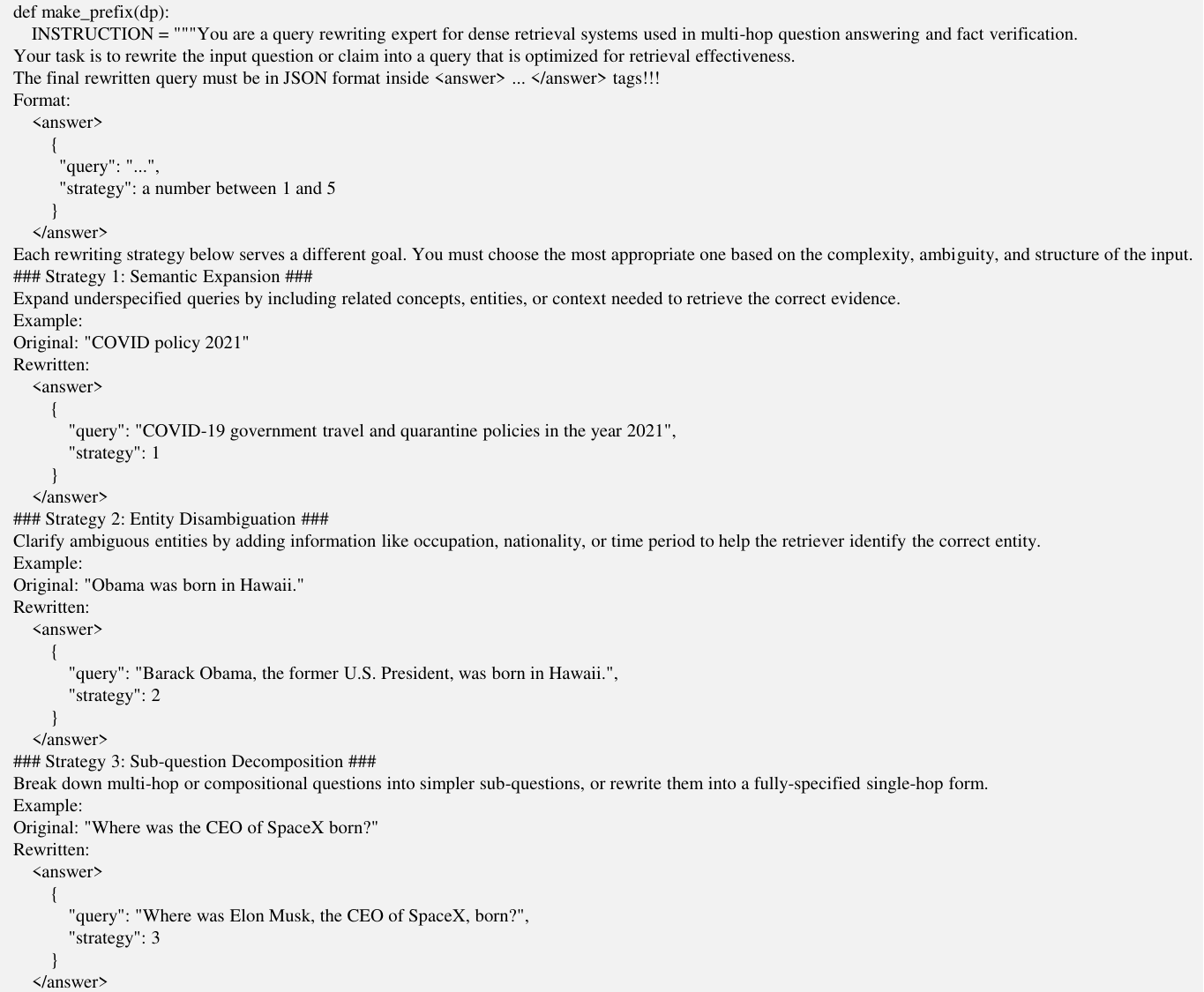}
\includegraphics[width=0.9\textwidth]{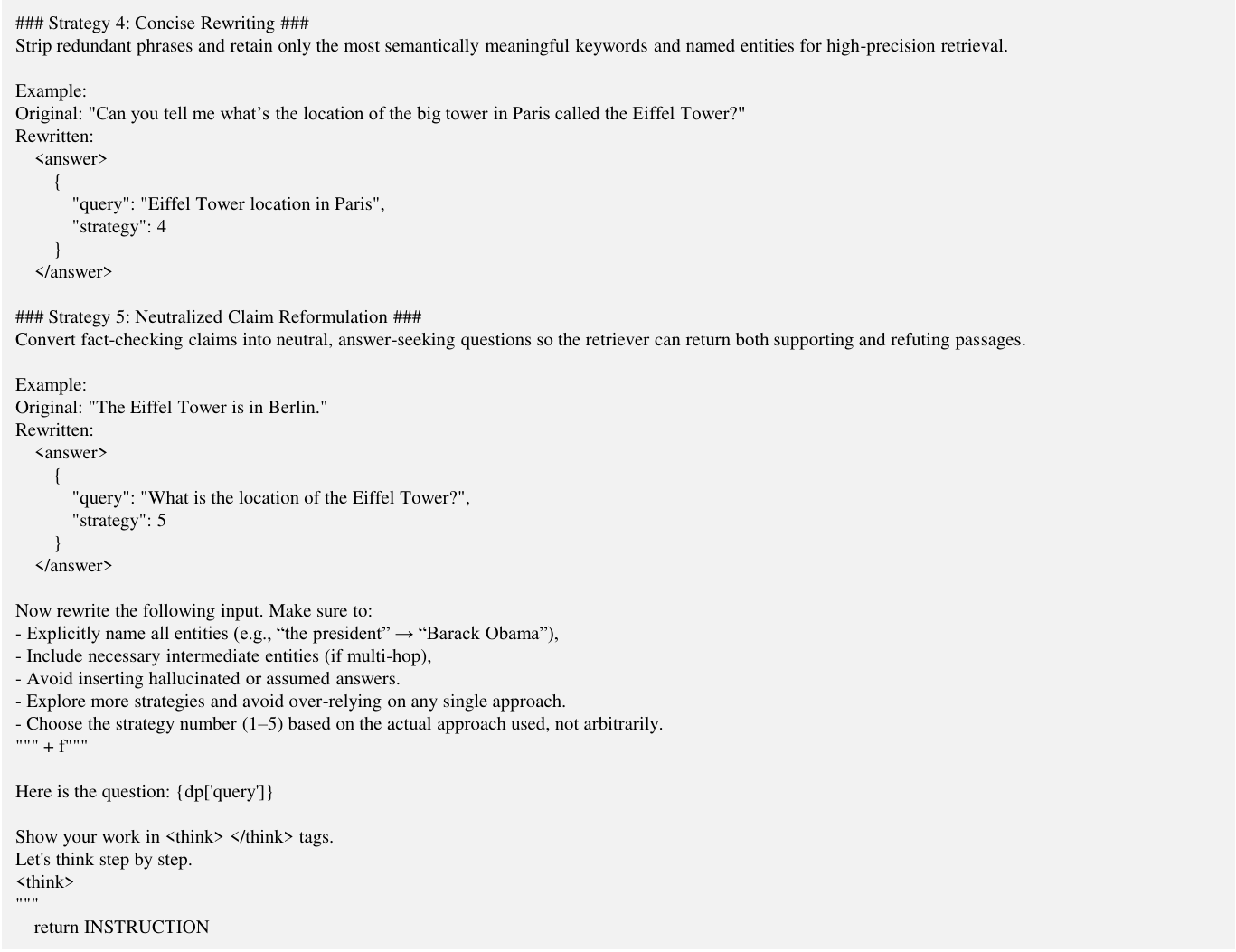}

\caption{Full Prompt Structure for Strategy-Guided Rewriting. The figure presents the complete instructional prompt provided to the LLM agent. It specifies the overall task, formally defines each of our five expert-crafted strategies, and provides illustrative examples to guide their application.}
\label{fig:strategy}
\end{figure*}

As referenced in Section~\ref{sec:method:strategy}, we present detailed strategies and the complete prompt used to guide our policy model in Table~\ref{tab:strategies} and Figure~\ref{fig:strategy}.

\subsection{SGP Performance with Contriever}

\label{appendix:contriever_scaling_law}

\begin{figure*}[t]
\centering
\includegraphics[width=1.0\textwidth]{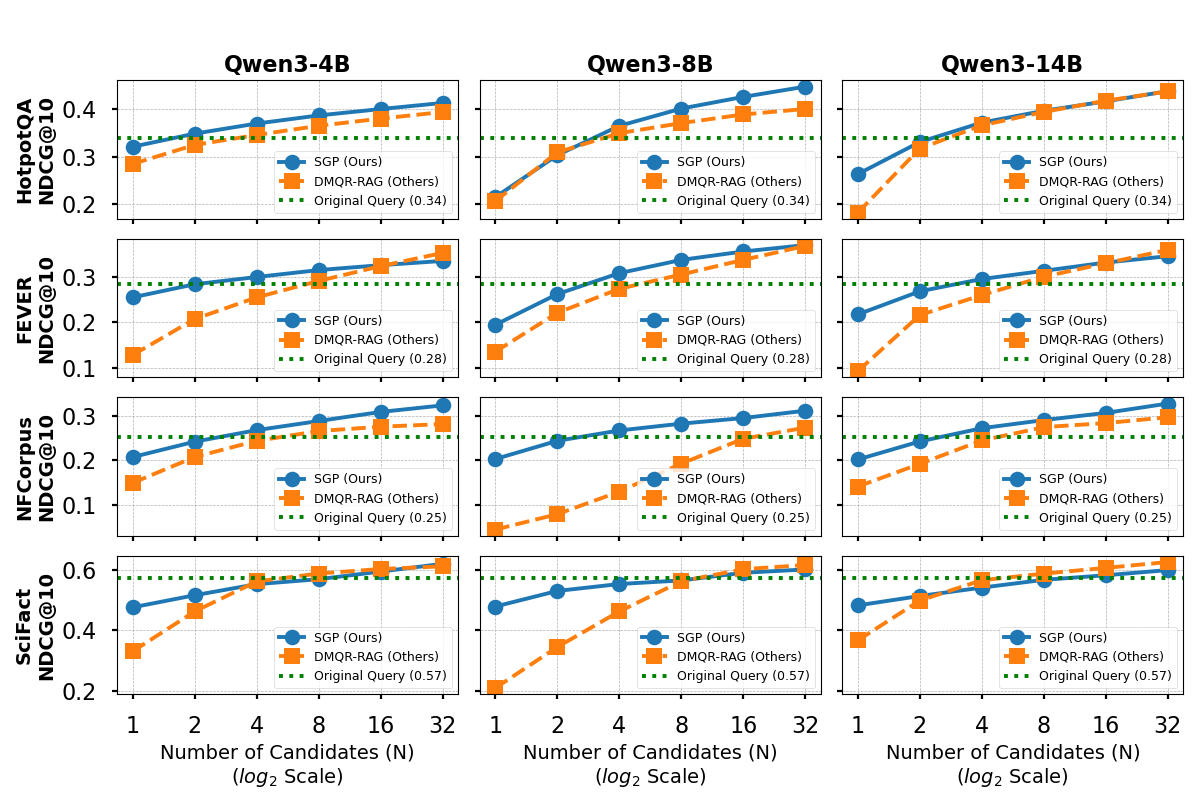}
\caption{Performance scaling laws for query rewriting using the Contriever~\cite{contriever} retriever across four benchmark datasets: HotpotQA~\cite{hotpotqa}, FEVER~\cite{fever}, NFCorpus~\cite{nfcorpus}, and SciFact~\cite{scifact}. Each subplot compares the upper-bound performance (NDCG@10) of our SGP against the strategies from \citet{DMQR-RAG}, evaluated via a Best-of-N methodology across the Qwen3 model series~\cite{qwen3}. The results demonstrate the superior scaling potential and higher performance ceiling of our approach.}
\label{fig:contriever_scaling_law}
\end{figure*}

To confirm that our findings are robust and not specific to a single retriever, we replicated the experiment using Contriever~\cite{contriever}. As detailed in Fig~\ref{fig:contriever_scaling_law}, our strategy again demonstrates a significant performance margin over the methods proposed by DMQR-RAG\cite{DMQR-RAG}.

\subsection{Training Hyperparameters}
\label{appendix:hyper}
For all experiments, we fine-tune the Qwen3-4B~\cite{qwen3} model using Verl~\cite{hybridflow} and GROP~\cite{grpo}, a reinforcement learning framework. We experiment with rollout numbers of 16 and 32. All training is conducted on 8 NVIDIA A100 GPUs. Unless otherwise specified, we use the default hyperparameters provided by Verl.

\subsection{Detailed Training and Response Length Curves}
\label{appendix:training_dynamics}
\begin{figure*}[h]
\centering
\includegraphics[width=1.0\textwidth]{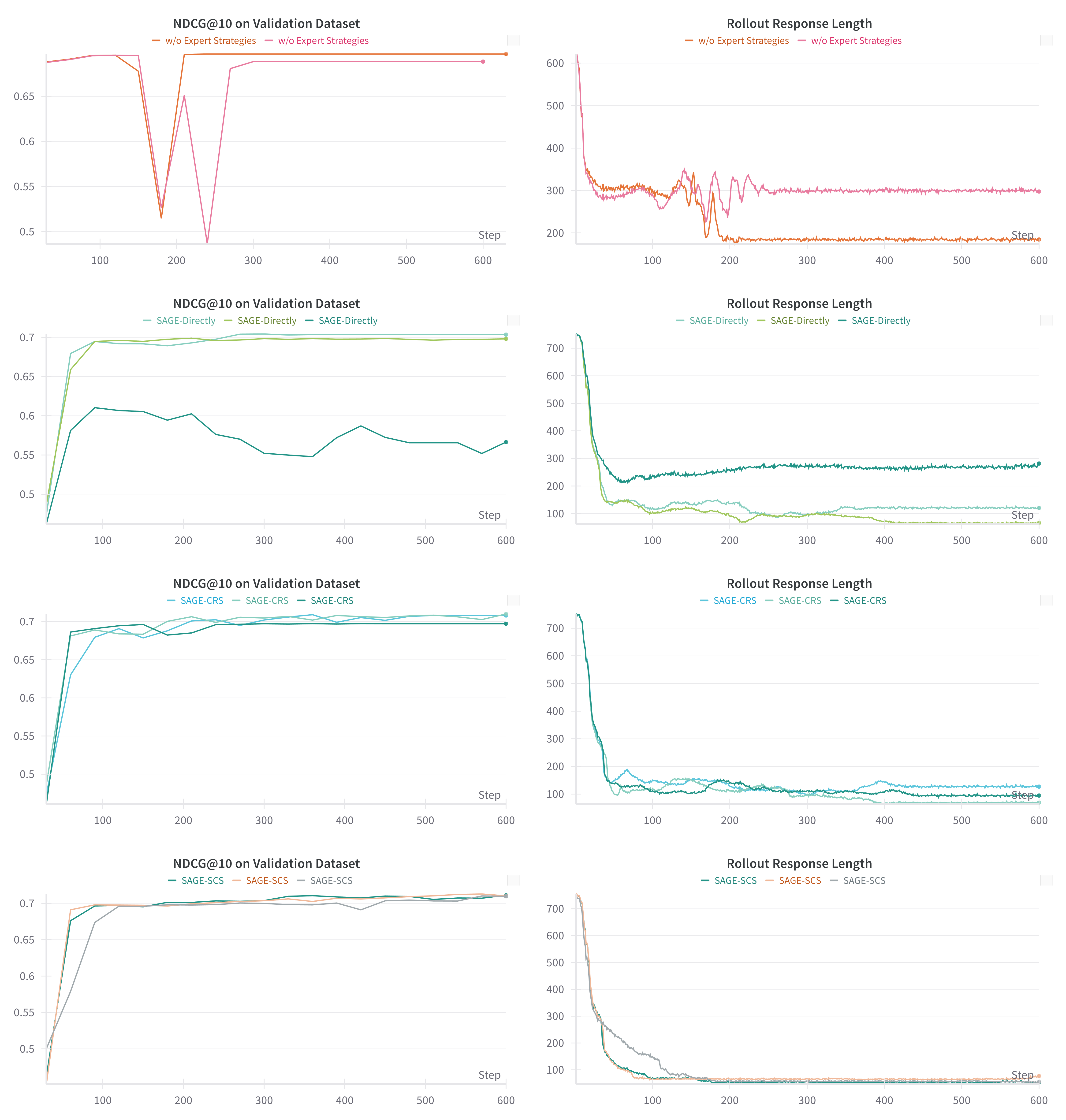}
\caption{Training dynamics on the HotpotQA dataset. Each row corresponds to a different method: (from top to bottom) baseline without strategies, SAGE (Direct), SAGE-CRS, and SAGE-SCS. The left column plots the validation NDCG@10   (evaluated every 30 steps) over training steps, while the right column plots the average response length. The baseline stagnates early, whereas all SAGE variants show continuous improvement in NDCG@10 and a steady decrease in response length.}
\label{fig:hotpotqa_training_dynamics}
\end{figure*}

\begin{figure*}[h]
\centering
\includegraphics[width=1.0\textwidth]{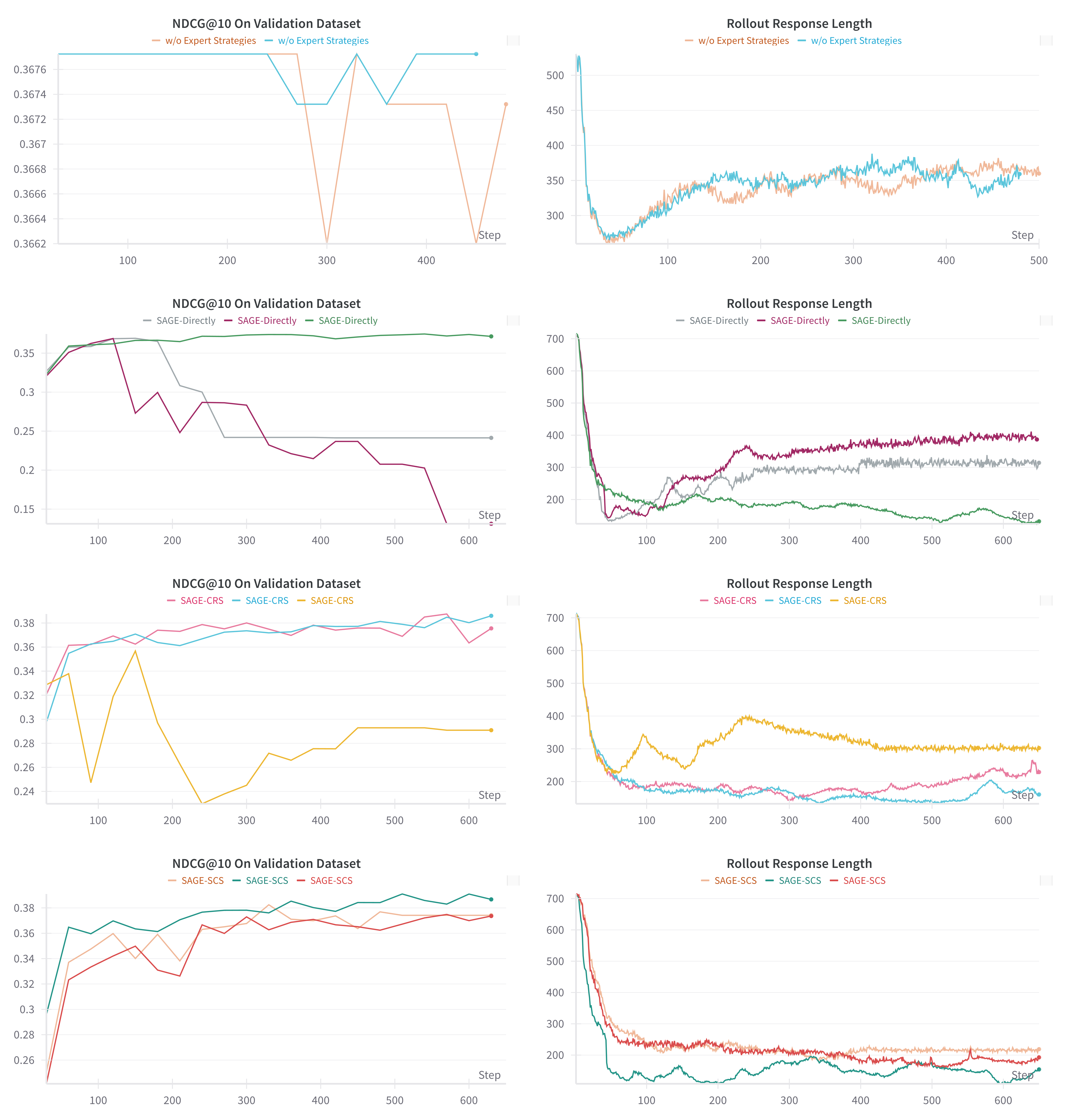}
\caption{Training dynamics on the NFCorpus dataset. Each row corresponds to a different method. The left column plots validation NDCG@10(evaluated every 30 steps), and the right column shows average response length. As with HotpotQA, the baseline model's performance plateaus while exhibiting high response length. In contrast, our SAGE variants, particularly SAGE-SCS and SAGE-CRS, demonstrate both stable performance gains and a significant reduction in response length over the course of training.}
\label{fig:nfcorpus_training_dynamics}
\end{figure*}

This section provides the detailed learning curves that support our analysis of training dynamics in the main text. Fig.~\ref{fig:hotpotqa_training_dynamics} and Fig.~\ref{fig:nfcorpus_training_dynamics} illustrate the validation NDCG@10 and the average training rollout response length as a function of training steps. These plots compare the learning trajectories of our SAGE variants against a baseline model trained without strategies. The results demonstrate that while the baseline model's performance quickly stagnates, our SAGE framework exhibits a superior learning trajectory, achieving continuous improvement in effectiveness while simultaneously learning to generate more concise responses.

\subsection{Related Work}
\paragraph{The Evolution of Query Rewriting}
Query rewriting has long been a cornerstone of IR, aimed at bridging the semantic gap between user intent and document representations. Early approaches typically relies on rule-based methods, thesaurus expansion, or statistical machine translation techniques. Although effective in specific contexts, these approaches often lack robustness and required extensive domain-specific feature engineering or sizable parallel corpora~\cite{query_aug, traditional_query_aug1, traditional_query_aug2, traditional_query_aug3, traditional_query_aug4, traditional_query_aug5}.
The advent of pre-trained sequence-to-sequence models~\cite{roberta,bert}, such as BART~\cite{bart} and T5~\cite{t5}, significantly reshape query rewriting, framing it explicitly as a supervised fine-tuning task.
However, this new paradigm introduces a critical bottleneck: heavy reliance on large-scale, high-quality annotated query pairs, which are costly and labor-intensive to construct.

\paragraph{Reinforcement Learning for Query Rewriting}
To alleviate reliance on explicit supervision, recent approaches leverage RL to guide the generative capabilities of LLMs using weaker signals~\cite{ppo, grpo, deep_retrieval, DMQR-RAG, HRM,MaFeRw, bpp-search}. 
However, the application of these powerful algorithms often reveals significant limitations. A common approach is to treat the LLM as a black box, applying on-policy algorithms like PPO without modification~\cite{deep_retrieval}, which overlooks the need for specialized guidance. Other lines of work focus on prompt engineering, but strategies tailored for sparse retrieval in general web search~\cite{DMQR-RAG} exhibit limited effectiveness when transferred to dense retrieval tasks. Furthermore, even approaches that integrate reward models, such as \citet{MaFeRw}, can be flawed; they often rely on an arbitrary, static fusion of scores from multiple fine-tuned reward models, lacking dynamic feedback from the actual retrieval environment. 
This highlights a fundamental gap: existing RL-based methods frequently neglect the nuanced, domain-specific guidance required to fully exploit the potential of dense retrieval systems.
In contrast, our SAGE framework addresses this gap by introducing a novel layer of explicit, human-designed strategic guidance. It achieves this through two core innovations: first, by equipping the agent with a set of fine-grained strategies for query augmentation, and second, by proposing novel reward calculation schemes like SCS and CRS to provide a more nuanced and effective learning signal.

\paragraph{Challenge in RL: Reward Hacking}
Beyond the lack of strategic guidance, applying RL to generative tasks introduces a further challenge. A pervasive issue in reinforcement learning is reward hacking, where an agent learns to maximize a misspecified or ambiguous reward function in unintended ways. This often results in the agent securing high scores by exploiting loopholes, rather than mastering the desired behavior or accomplishing the true underlying objective~\cite{lilianweng,reward_hacking_in_llm1,reward_hacking_in_llm2,leverageLLMMIP}.

In the context of query rewriting, this problem manifests in a particularly deceptive form. Since the policy model has access to the original query, it can discover a trivial, low-effort strategy: leave the query unchanged. Given that modern retrievers can achieve high baseline performance on the original queries for many benchmarks, this "do nothing" policy is often reinforced with a strong, positive reward. This effectively traps the agent in a local optimum, allowing it to "hack" the reward by avoiding the risk of exploration altogether. Our work directly confronts this challenge by introducing mechanisms to force meaningful exploration.

\end{document}